\PassOptionsToPackage{numbers}{natbib}  % 必须在 \documentclass 之前
\documentclass{article}
\usepackage{svg}
\usepackage{fontawesome}
\usepackage{pifont}
\usepackage{graphicx}
\usepackage{wrapfig}
\usepackage{xcolor}
\usepackage{pifont}
\usepackage{multirow}
\definecolor{grayblue}{RGB}{170, 180, 230}
\definecolor{lightblue}{RGB}{150, 200, 255}
\usepackage[colorlinks=true, citecolor=grayblue]{hyperref}
\newcommand{\xmark}{\ding{55}}%

\usepackage[preprint]{neurips_2025}

\usepackage[utf8]{inputenc} % allow utf-8 input
\usepackage[T1]{fontenc}    % use 8-bit T1 fonts
\usepackage{hyperref}       % hyperlinks
\usepackage{url}            % simple URL typesetting
\usepackage{booktabs}       % professional-quality tables
\usepackage{amsfonts}       % blackboard math symbols
\usepackage{nicefrac}       % compact symbols for 1/2, etc.
\usepackage{microtype}      % microtypography
\usepackage{amsmath} 

\title{Multi-human Interactive Talking Dataset}

\author{%
    Zeyu Zhu, Weijia Wu, Mike Zheng Shou$^{({\textrm{\faEnvelopeO}})}$ \\
    \\
    Show Lab, National University of Singapore \\
}

\begin{document}

\maketitle
\let\thefootnote\relax\footnotetext{$^{\textrm{\faEnvelopeO}}$ Corresponding author.}

\begin{abstract}
Existing studies on talking video generation have predominantly focused on single-person monologues or isolated facial animations, limiting their applicability to realistic multi-human interactions.
To bridge this gap, we introduce MIT, a large-scale dataset specifically designed for multi-human talking video generation.
To this end, we develop an automatic pipeline that collects and annotates multi-person conversational videos.
The resulting dataset comprises 12 hours of high-resolution footage, each featuring two to four speakers, with fine-grained annotations of body poses and speech interactions.
It captures natural conversational dynamics in multi-speaker scenario, offering a rich resource for studying interactive visual behaviors.
To demonstrate the potential of MIT, we furthur propose CovOG, a baseline model for this novel task.
It integrates a Multi-Human Pose Encoder (MPE) to handle varying numbers of speakers by aggregating individual pose embeddings, and an Interactive Audio Driver (IAD) to modulate head dynamics based on speaker-specific audio features.
Together, these components showcase the feasibility and challenges of generating realistic multi-human talking videos, establishing MIT as a valuable benchmark for future research. The code is avalibale at: \url{https://github.com/showlab/Multi-human-Talking-Video-Dataset}
\end{abstract}

\section{Introduction}

Recent advancements in human-centric video generation~\cite{li2024openhumanvid,lei2024comprehensive} have markedly improved the synthesis of high-fidelity human videos. Among the most prominent research directions are pose-guided animation~\cite{chang2023magicdance,ma2024follow,hu2024animate,xu2024magicanimate}, which enables fine-grained control over full-body movements, and audio-driven talking avatar generation~\cite{chen2024echomimic,cui2024hallo2,zhu2023taming}, which focuses on producing accurate lip synchronization and expressive head motion conditioned on speech. Within the domain of audio-driven generation, substantial progress has been made in co-speech gesture synthesis~\cite{ginosar2019learning} and talking head animation~\cite{son2017lip, wang2023agentavatar}. The former seeks to align upper-body gestures with spoken content, while the latter aims to generate realistic facial expressions, head poses, and lip movements driven by audio input, thereby enhancing the expressiveness and naturalness of talking avatars. Despite these advances, existing methods predominantly focus on \textit{single-person monologues} or \textit{isolated facial regions}, lacking the capacity to model multi-speaker interactions. This limitation significantly constrains their applicability in realistic settings such as interviews, panel discussions, or films, where natural, multi-party conversations are essential. 

In contrast to single-speaker scenarios, multi-speaker interactions involve complex dynamics, including turn-taking, fluid role transitions between speaking and listening, and non-verbal communicative behaviors such as eye contact and gesturing. Moreover, current datasets~\cite{chung2018voxceleb2, ginosar2019learning} and generation frameworks~\cite{liu2024customlistener,liu2024tango} are not designed to capture such multi-speaker conversational dynamics. Although recent work such as INFP~\cite{zhu2024infp} has taken initial steps toward interactive talking-head generation with multiple speakers, it remains restricted to facial animation alone. As a result, it fails to incorporate full-body behavioral cues critical for modeling realistic social interactions, thereby limiting both the quality and application of the generated content.

\begin{figure}[t]
    \centering
    \includegraphics[width=1\linewidth]{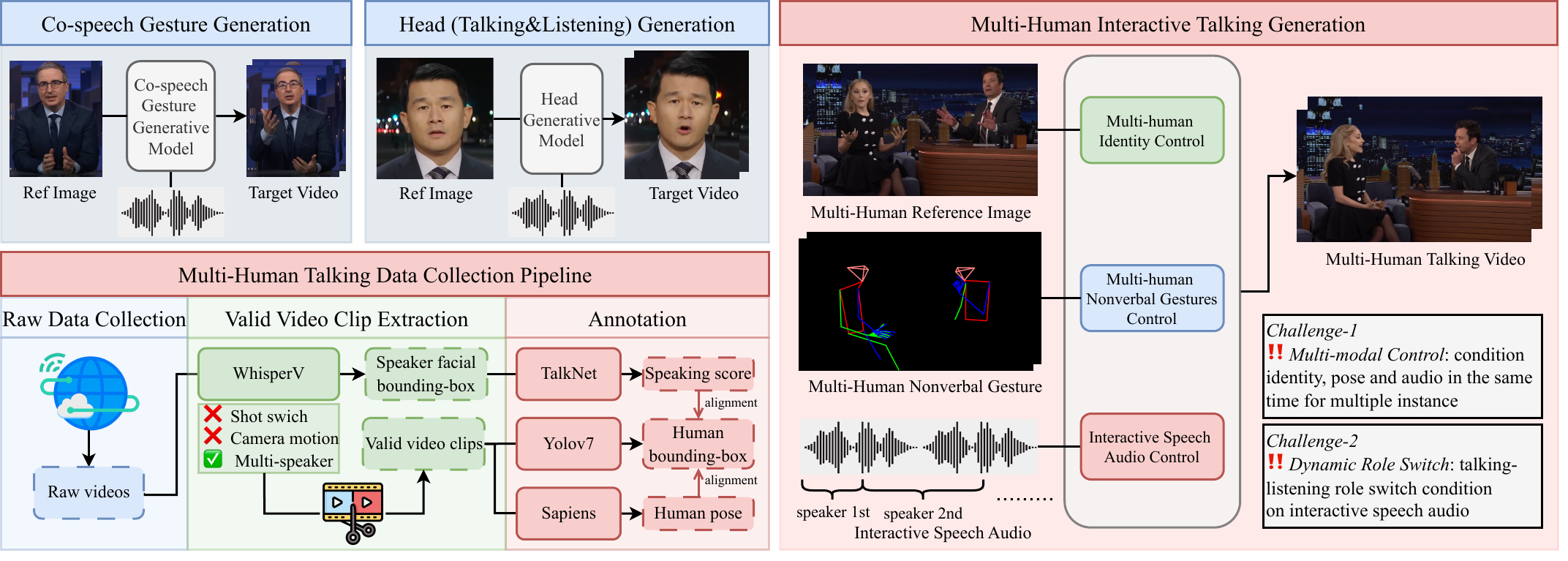}
    \caption{\textbf{Single Speaker Generation} \textit{v.s.} \textbf{Mulit-human Interactive Talking Generation} and \textbf{Automatic Data Collection Pipeline}.
    The pipeline of existing tasks are shown in \textcolor{lightblue}{blue}, Co-speech Gesture Generation~\cite{ginosar2019learning, liu2024tango}, and Talking or Listening Head Generation~\cite{cui2024hallo2, wang2023agentavatar}. 
    In contrast, Multi-person Interactive Talking Generation enables dynamic speaker interactions by incorporating identity, interactive pose and audio control, as shown in \textcolor{pink}{red}. 
    %In contrast, multi-human talking generation enables rich, natural, and dynamic interactions among multiple speakers in a conversation, enhancing realism and engagement.
    %
    And the automatic data collection is shown consisting of raw data collection, valid video clip extraction and annotation.
    }
    \label{fig:motivation}
\end{figure}

To advance beyond the limitations of single-speaker and facial-only generation, we define a new task, Multi-Human Talking Video Generation, which aims to synthesize realistic multi-person talking videos conditioned on reference images, body poses, and speech audio, as illustrated in Figure~\ref{fig:motivation}. Constructing a dataset suitable for this task is particularly challenging, as it requires the accurate extraction of multi-person conversational scenes, stabilization of camera motion, and the removal of occlusions and post-production artifacts. In this paper, we propose an automatic data collection pipeline and use it to build a benchmark for this task. % collect pipeline
Specifically, we introduce the Multi-human Interactive Talking dataset(MIT), a fine-grained collection of 12 hours of multi-human videos featuring 2--4 speakers with diverse identities. This dataset includes multi-human pose annotations aligned with each speaker's speaking score label that indicates whether the human is speaking. Furthermore, we propose a baseline model designed for this task, namely CovOG: ConversationOriginal. Built on AnimateAnyone~\cite{Moore-AnimateAnyone}, CovOG integrates two key components: the Multi-Human Pose Encoder (MPE) and the Interactive Audio Driver (IAD). The MPE aggregates individual pose embeddings, allowing the model to accommodate a flexible number of human speakers. Meanwhile, the IAD dynamically refines speaker-specific head and pose features using an audio-driven speaking score, ensuring smooth and natural transitions between speaking and listening. Our work aims to lift audio-driven human-centric video generation to a more realistic setting, offering a significant contribution to the field. 

% To advance beyond, we explore and define a new task: Multi-Human Talking Video Generation, as shown in Figure~\ref{fig:motivation}, which generate \textbf{multi-human} talking videos condition on reference image, pose gestures and speech audio.
% %
% Constructing a multi-person talking dataset is particularly demanding, requiring precise extraction of multi-human conversation shots, stabilization of camera motion, and removal of occlusions and post-editing artifacts. 
%
% Meanwhile, existing methods struggle with effectively controlling multimodal features and capturing the dynamic nature of speech interactions, further complicating their application to this task.

% In this paper, we propose a novel dataset and a well-designed model for specially multi-human video generation. 
%

% %
% %
% Furthermore, we propose a novel model designed for this task, namely CovOG: ConversationOriginal. 
% %
% Built on AnimateAnyone~\cite{Moore-AnimateAnyone}, CovOG integrates two key components: the Multi-Human Pose Encoder (MPE) and the Interactive Audio Driver (IAD). 
% %
% The MPE aggregates individual pose embeddings, allowing the model to accommodate a flexible number of human speakers. 
% %
% Meanwhile, the IAD dynamically refines speaker-specific head and pose features using an audio-driven speaking score, ensuring smooth and natural transitions between speaking and listening.

To summarize, the contributions of this paper are:
\begin{itemize}
\item To the best of our knowledge, we first explore multi-human talking generation which lift exiting audio-driven video generation to a more realistic, universal setting.

\item We develop an automatic data collection pipeline and construct the first dataset for multi-human talking video generation, featuring annotations of pose and speech interaction.

\item We present a baseline model for this novel task, which supports a flexible number of human speakers and captures the dynamics of speech interactions. We further conduct extensive studies to benchmark our baseline against existing methods and analyze its performance.

\end{itemize}

\section{Related Work}
\subsection{Human-centric Video Generation Model}
Recent advancements in diffusion models~\cite{rombach2022high,singer2022make,guo2023animatediff,chen2023videocrafter1,zhang2024show} have significantly enhanced video generation in terms of length, quality, and controllability. Stable Video Diffusion~\cite{blattmann2023stable} employs latent diffusion to model video distributions within a latent space, enabling efficient and high-quality video synthesis. Furthermore, DiT-based models~\cite{peebles2023scalable}, such as CogVideoX~\cite{yang2024cogvideox} and MovieGen~\cite{sepehri2024mediconfusion}, improve video length and fidelity by diffusion transformers. Building on the advancements of these base models, human-centric video generation~\cite{li2024openhumanvid,lei2024comprehensive} has garnered increasing attention due to its significant application potential. Text-driven models, such as Performer ~\cite{jiang2023text2performer} and DirectorLLM~\cite{song2024directorllm}, synthesize diverse human motions based on text prompts. Meanwhile, pose-based methods~\cite{feng2023dreamoving,chang2023magicdance,ma2024follow} generate fine-grained controllable motions by leveraging pose sequences and reference images. Notably, AnimateAnyone~\cite{hu2024animate} employs ControlNet~\cite{zhang2023adding} to maintain identity consistency throughout motion synthesis, while MagicAnimate~\cite{xu2024magicanimate} integrates an additional control branch to achieve better pose alignment. 
%These advancements collectively contribute to the development of more realistic, controllable, and high-fidelity human-centric video generation techniques, expanding their applicability across various domains.

\subsection{Audio-Driven Character Animation}
\textbf{Single Portrait Image Animation.} Single portrait image animation, which generates a talking or listening head from a given audio and portrait image, has recently gained significant attention. In talking head generation, various datasets~\cite{son2017lip, chung2018voxceleb2, sung2024multitalk} have been proposed. Notably, MEAD~\cite{wang2020mead} focuses on emotion control, offering data across eight emotions with three intensity levels, while CelebV-HQ~\cite{zhu2022celebv} provides diverse identities in realistic settings. Early approaches~\cite{prajwal2020lip, vougioukas2020realistic, zhang2023sadtalker} relied on GAN-based models to improve lip synchronization. Recently, diffusion-based models~\cite{stypulkowski2024diffused, jiang2024loopy, chen2024echomimic, cui2024hallo2, wang2024emotivetalk} have significantly enhanced realism, consistency, and control ability. In listening head modeling, RLHG~\cite{zhou2022responsive} first proposed ViCo dataset and built a sequential auto-encoder to generate non-verbal facial feedbacks given the speech audio and portrait image. Recent approaches~\cite{huang2022perceptual, ng2022learning, geng2023affective, liu2024customlistener} have advanced reaction quality and controllability(\textit{e.g.}, pose and text), by leveraging superior generative models(\textit{e.g.}, VQ-VAE) and LLMs.

\noindent\textbf{Single-human Co-speech Generation.}
Co-speech generation enhances single-head generation by incorporating nonverbal gestures, making the content more expressive. To facilitate research in this area, a high-quality dataset, SSGD~\cite{ginosar2019learning}, has been developed, providing co-speech video clips of 10 speakers along with pose annotations. Early approaches~\cite{qian2021speech, liu2022learning, zhu2023taming, he2024co} typically follow a two-stage pipeline: first, human poses are generated based on speech audio, and subsequently, pose-to-video methods (\textit{e.g.}, AnimateAnyone~\cite{hu2024animate}) are employed to synthesize co-speech gesture videos using a reference image. More recently, some studies have explored retrieval-based solutions for this task. Gesture video reenactment~\cite{zhou2022audio, liu2024tango} utilizes a short reference video clip (\textit{e.g.},, two minutes) to generate stylized gesture videos that align with novel speech inputs, resulting in more faithful and visually coherent outputs.

\noindent\textbf{Multi-human Conversation Generation.}
Despite notable advancements in audio-driven single-human animation, it remains limited in capturing the richness of multi-human interactive conversations, which are more common and expressive in real-world applications (\textit{e.g.}, movie dialogues, talk show interviews, and live streams). Recently, several studies~\cite{wang2023agentavatar, zhou2023interactive, tran2024dyadic} have explored interactive head generation, producing two talking-listening heads in a dyadic manner forming a conversation. Notably, INFP~\cite{zhu2024infp} introduced a large-scale dataset comprising extensive head-only conversational videos between two individuals and proposed an interactive motion guide to facilitate seamless talking-listening transitions. These approaches are constrained to generate only two individuals' head areas, as they fail to incorporate non-verbal contents such as eye contact, physical interaction, thereby restricting their applicability in more dynamic and natural conversational full-body interaction settings. Moreover, existing studies primarily focus on ideal turn-taking scenarios, where speakers alternate systematically, while challenges such as rapid role-switching and overlapping speech remain inadequately addressed. Existing methods fail to address multi-human talking generation in terms of full-body interactions and dynamic talking patterns, which requires specific models and datasets to capture multi-human interactive talking videos.

\begin{figure*}[htbp]
    \centering
    \includegraphics[width=\textwidth]{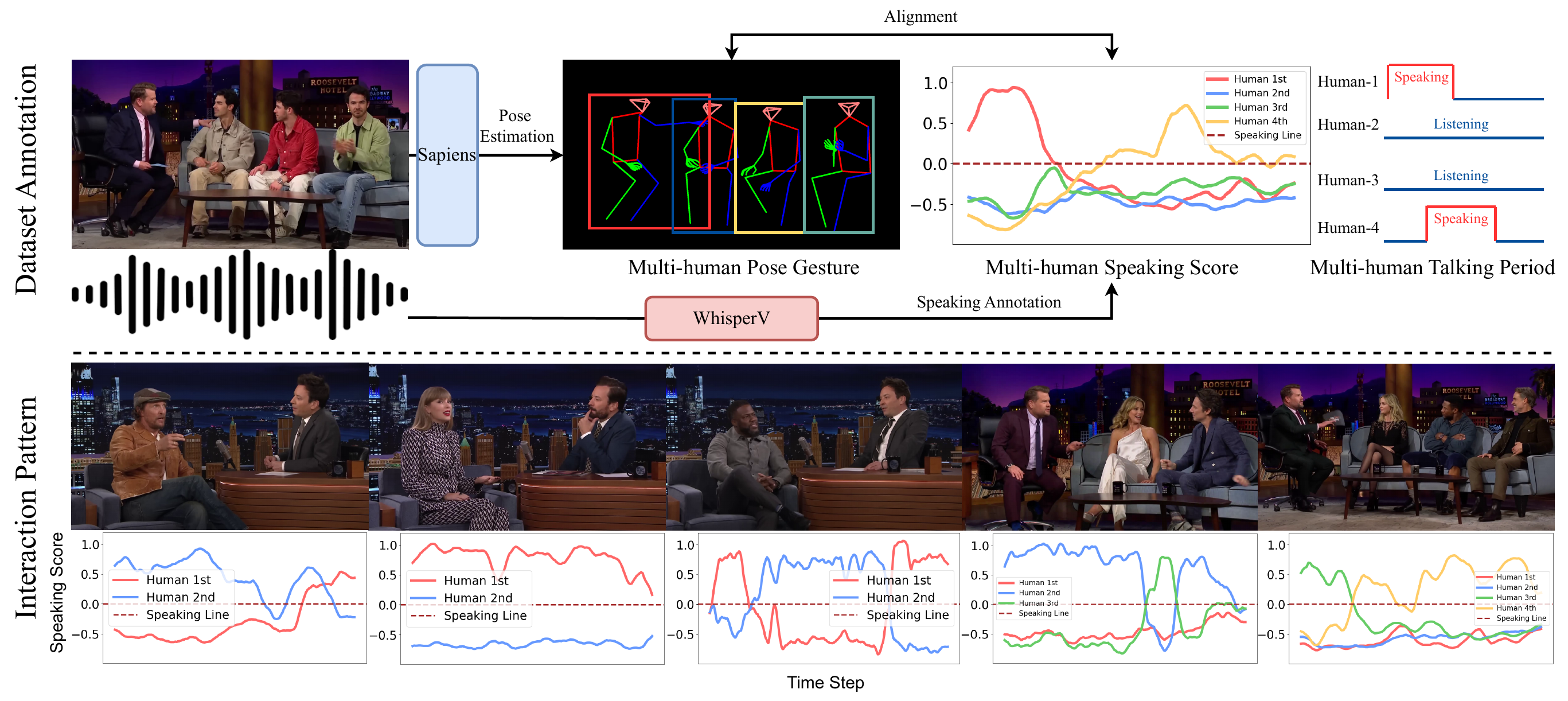}  
    \caption{\textbf{ Multi-human Interactive Talking Dataset.} 
    Sapiens~\cite{khirodkar2024sapiens} and WhisperV~\cite{whisperV} are used to annotate multi-human gesture and interactive speech respectively.
    %and align them with bounding box obtained by Yolov7\cite{wang2023yolov7}.
    %
    % 2) 
    MIT dataset captures rich conversation interaction pattens of multi-human, such as talking-listening, tune-talking, over-talking and other complex pattens.
    }
    \label{fig:datasets}
\end{figure*}
\section{Multi-Human Interactive Talking Datset}
\label{sec:data}
We present a high-quality dataset for multi-human interactive talking video generation, comprising over $12$ hours of high-resolution conversational clips with diverse interaction patterns and approximately $200$ distinct identities.
The dataset was constructed through a fully automated pipeline, facilitating future scale-up with minimal manual intervention. We provide a detailed description of this process in the following subsections, covering the data collection methodology (Section~\ref{datacollection}) and a analysis of interaction types and annotation statistics (Section~\ref{dataAnnotation}).

%
% As illustrated in Figure~\ref{fig:datasets}, our MIT dataset contains conversation video clips involving $2$ to $4$ individuals, with annotations for human poses and speaking interactions, where the speaking annotation indicates whether a participant is actively speaking at the given moment. 
%
% In this section, we provide a detailed overview of the dataset, including data collection~(%Sec.%
% \S\ref{datacollection}) and annotation process~(%Sec.%
% \S\ref{dataAnnotation}).
% Notably, our data collection and processing pipeline is fully automated, enabling future scale-up. 
%
%Further details are as below. 

\subsection{Automatic Data Collection Pipeline}
\label{datacollection}
As illustrated in Figure~\ref{fig:motivation}, the data collection pipeline comprises three main stages: raw video collection, valid clip extraction, and multi-modal annotation. First, conversational videos are collected from online platforms. However, most real-world videos undergo post-editing and include multiple shots from different perspectives (\textit{e.g.}, close-up shots of faces and wide shots of the entire scene), which are unsuitable for current video generation models that require temporally consistent visual content. To address this, WhisperV~\cite{whisperV} is adopted to segment videos into individual shots and to track facial trajectories of speakers within each shot. Clips featuring multiple active speakers within a single continuous shot are then extracted to preserve interactive dynamics. Finally, foundational perception models are employed to extract speaking scores, human poses, and bounding boxes. The bounding boxes serve as spatial anchors to align multi-modal signals, enabling consistent annotation for each individual speaker.

\noindent\textbf{Pose Annotation.} As part of the annotation process, 2D skeletal keypoints are extracted using Sapiens-2B~\cite{khirodkar2024sapiens} in the COCO133~\cite{jin2020whole} format.  
A subset of $59$ keypoints is selected to represent the head, body, arms, legs, and hands, as illustrated in Figure~\ref{fig:datasets}.  
Specifically, only three keypoints are retained for the head to define its orientation, as finer facial expressions (e.g., lip movement, emotions) are primarily driven by audio. Notably, although the detected pose keypoints are pseudo-labels rather than manually annotated ground truth, they are obtained using a state-of-the-art pose estimation model, similar to SSGD~\cite{ginosar2019learning}.
This provides sufficient accuracy for generation tasks despite the absence of human supervision.

\noindent\textbf{Speaking Score.}  In parallel, speaking scores are extracted using TalkNet~\cite{beliaev2020talknet}, a model that performs speech activity detection. As illustrated in Figure~\ref{fig:datasets}, each individual is associated with a speaking score curve indicating periods of speech and silence.  
A score approaching 1 indicates active speaking, while a score nearing -1 corresponds to non-speaking states. The figure further illustrates how speaking scores reflect various interaction patterns: clear alternation between high and low scores indicates speaker turns; overlapping high scores across speakers correspond to simultaneous speech; and smooth transitions between high and low values capture speaking–listening dynamics.

\noindent\textbf{Pose–Speech Alignment.} 
After obtaining pose annotations and speaking scores—which are independently extracted and thus not inherently aligned—alignment is performed for each individual using human bounding boxes detected by YOLOv7. For each frame, pose annotations are assigned to the individual whose bounding box contains the highest number of keypoints. Similarly, each face track is matched to the individual whose bounding box most frequently overlaps with the facial bounding boxes across frames, leveraging the fact that face tracks are already aligned with speaking scores. By using the human bounding box as a shared spatial reference, both pose and speech annotations are consistently associated with the correct individual.

% %
% As shown in Figure~\ref{fig:datasets}, we first collect full video and process them using WhisperV~\cite{whisperV}.
% %
% Since our focus is on shots featuring multiple speakers, we employ face detection to track faces within each shot.
% %
% Then we extract clips of at least $5$ seconds but no more than $10$ seconds($25$ fps) that contain more than two individuals by getting the intersection of face tracks.
% %
% We further employ YOLOv7~\cite{wang2023yolov7, RizwanMunawaryolov7} for human detection, filtering out cases that only contain head regions or atypical clips that do not resemble conversational videos.
% %
% As camera motion is very rare within the single shot from these two channels, no additional filtering is required for this.

\subsection{Dataset Analysis}
\label{dataAnnotation}
\noindent\textbf{Data Source.} Real-world videos often contain camera motion, occlusions, and post-editing artifacts, which are challenging to remove and typically require extensive manual intervention, such as region-specific inpainting. To mitigate these issues while ensuring diverse and interactive multi-speaker scenarios, we curate classic and representative interview videos from two channels—\textit{The Tonight Show}\footnote{https://www.youtube.com/@fallontonight} and \textit{The Late Late Show}\footnote{https://www.youtube.com/@TheLateLateShow}—as our data sources. These videos feature interactive multi-speaker scenarios that reflect real-world social behaviors, captured with static camera setups and minimal occlusions, making them well-suited for training models on interactive talking video generation. Despite the limited scene variety, the dataset features complex interactions and diverse identities, demonstrating its potential applicability to news, live broadcasting, and cinematic content.

\noindent\textbf{Interaction Pattern.} 
Multi-human interaction patterns constitute a critical yet challenging aspect of generating talking videos with multiple speakers, due to their inherent diversity and complexity.  
The most common pattern is turn-taking, where speakers alternate their roles, as explored in prior works~\cite{zhu2024infp} for interactive talking head.  
However, real-world conversations often exhibit more intricate dynamics, such as interruptions (over-talking), pauses, and rapid shifts between speaking and listening roles.  
Figure~\ref{fig:datasets} illustrates the diverse interaction patterns captured in the MIT dataset, highlighting its suitability for advancing research in multi-human talking video generation.

\noindent\textbf{Dataset Statistics.} 
A comparison between MIT and existing datasets is presented in Table~\ref{tab:dataset_overview}.  
MIT is the only dataset that features multi-human full-body interactions within conversational contexts.  
Although the total duration is limited to 12 hours, the automated data collection pipeline enables future scalability, compensating for this limitation.

\noindent\textbf{Quality of Data Annotations.} On a subset of 20 testing videos, we evaluate the automatic pose detections against human annotations and find that the pseudo ground truth is sufficiently accurate for our task. We also manually annotate the speaking–listening transition points (\textit{i.e.}, the zero point of the speaking score) for each speaker, achieving an average temporal error below 0.1 second. Furthermore, we verify that pose–speaking alignments of all samples are correct.

\begin{table*}[t]
    \centering
    \caption{\textbf{Existing Datasets} \textit{v.s.} \textbf{MIT}. 
    Compared to previous datasets that focus on single-person speech and isolated facial animation, our MIT dataset uniquely features multi-person talking videos with full-body interactions.
    }
    \label{tab:dataset_overview}
    \begin{tabular}{lccccccc}
        \toprule
        \textbf{Dataset} & \textbf{Num.} & \textbf{Area} & \textbf{Character} & \textbf{Pose} & \textbf{Speak} & \textbf{Res.} & \textbf{Total Len.(h)}\\
        \midrule
        SSGD~\cite{ginosar2019learning} & One & Body & Speaking & $\checkmark$ &  \xmark & 1920×1080 & 144\\
        HDFTD~\cite{zhang2021flow} & One & Head & Speaking & \xmark &  \xmark & 512×512 & 16\\
        ViCo~\cite{zhou2022responsive} & One & Head & Listening & \xmark &  \xmark & 384×384 & 2\\
        RealTalk\cite{geng2023affective} & Two & Head & Interactive  & \xmark & $\checkmark$ & 1280x720 & 115\\
        DyConv~\cite{zhu2024infp} & Two & Head & Interactive & \xmark & $\checkmark$ & 400×400 & 200\\
        \midrule
        MIT & Multi & Body & Interactive& $\checkmark$ & $\checkmark$ & 1920×1080 & 12\\
        \bottomrule
    \end{tabular}
\end{table*}

\section{Baseline: CovOG}
% Despite recent progress, existing methods fall short in generating full-body interactive talking videos involving multiple speakers.  
% This is due to the complexity of modeling multi-person interactions under multi-modal control—specifically, reference images, pose sequences, and interactive speech audio for each individual.  
% The core challenge lies in effectively aligning these signals with their corresponding identities.  
To tackle this task, we introduce CovOG, a tailored model built upon the single-person animation framework AnimateAnyone~\cite{hu2024animate} which leverages Stable Diffusion~\cite{blattmann2023stable} as base model and ensures identity consistency through ReferenceNet while incorporating conditional poses by embedding their features into the latent space via Pose Guider. 
Expanding on this foundation, CovOG integrates two key modules: the Multi-Human Pose Encoder (\textit{i.e.}, Pose Guider/Adaptor) and the Interactive Audio Driver (IAD) as shown in Figure~\ref{fig:baseline}. 
% These modules are specifically designed to enable multi-human pose control and achieve facial region alignment with speech audio, as illustrated in Figure~\ref{fig:baseline}. 
The detail of each module is provided below.
% %
%along with multi-instance animation areas, including multiple speakers and the background.
%
% The key challenge lies in effectively conditioning the model to align each control signal with its corresponding animation area, ensuring precise and coherent motion synthesis and interaction modeling. To address these challenges, we introduce CovOG, a well-designed model built on the validated single-human animation framework, AnimateAnyone~\cite{hu2024animate}. 
%
% However, when applied to this task, three key challenges remain unsolved:

% \begin{itemize}
%     \item \textit{How to condition multi-human pose control with unfixed number of individuals?}
%     \item \textit{How to condition multi-human identity and background with a single reference image?}
%     \item \textit{How to condition interactive speech audio?}
% \end{itemize}

\subsection{Network Architecture}
\begin{figure*}[t]
    \centering
    \includegraphics[width=1\textwidth]{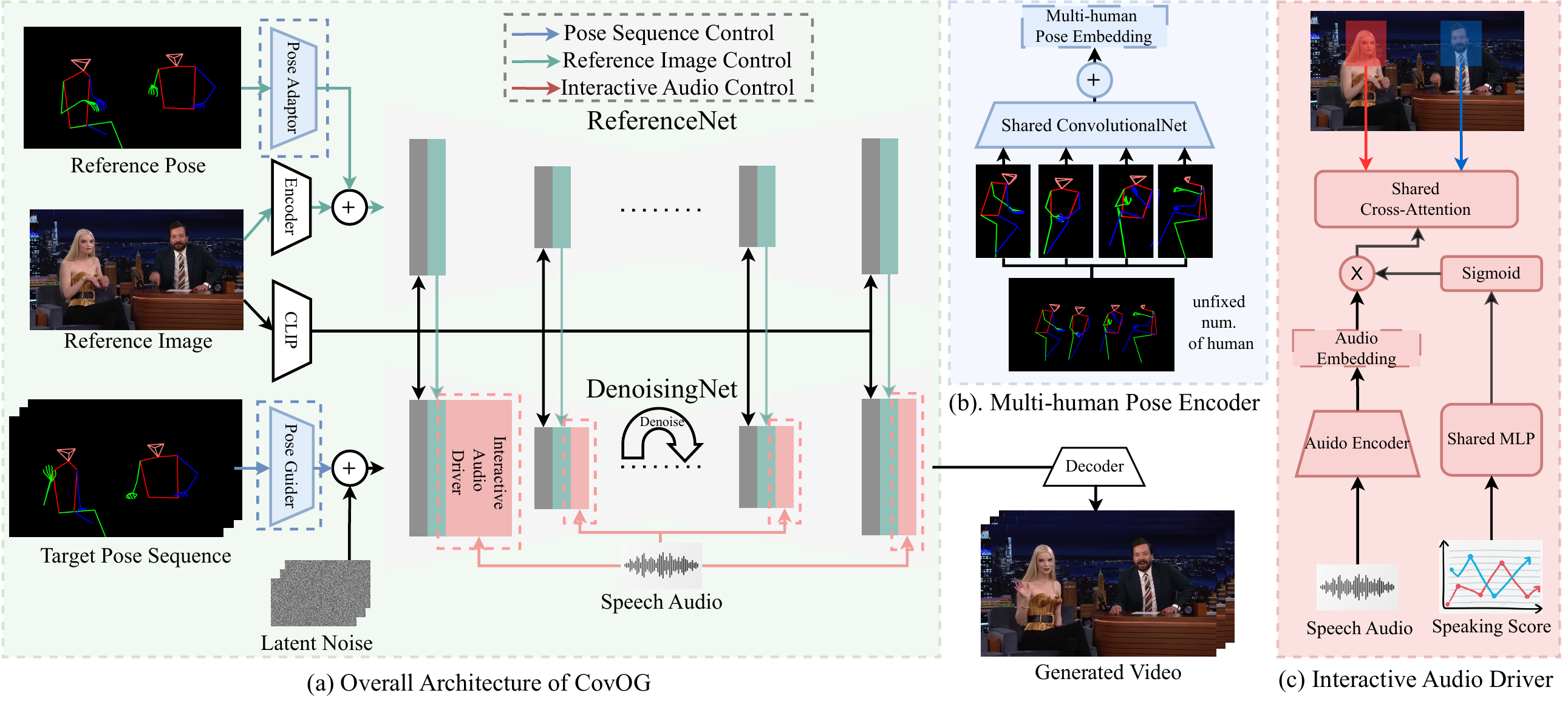}  
    \caption{\textbf{Overview of proposed method CovOG.}
    (a) The overall architecure of CovOG.
    (b) Implement of Multi-human Pose Encoder used in Pose Adaptor and Pose Guider.
    (c) Implement of Interactive Audio Driver to capture the dynamic facial interaction between multiple speakers.}
    \label{fig:baseline}
\end{figure*}

\noindent\textbf{Overview.} The overview of CovOG is shown in Figure~\ref{fig:baseline} (a). 
Specifically,
the multi-human pose embedding is incorporated into the multi-frame latent noise as pose control before being fed into DenoisingNet.
Additionally, ReferenceNet is introduced for identity control using reference images, while IAD modules are incorporated to control the facial area based on speech audio. 
%
% To recover the original latent representation, the DenoisingNet \( \epsilon_\theta \) is trained to predict the added noise component, thereby enabling the reconstruction of clean latent variables.
% %
% The training objective for learning the denoising process is defined as:
% \begin{align}
% L_t = \mathbb{E}_{z_0,r,p,s} \left[ \left\|\epsilon - \epsilon_\theta(\tilde{z},r,p,s,t)\right\|^2 \right],
% \end{align}
% where \( \epsilon_\theta \) represents the learnable network parameter, while \( r \), \( p \), and \( s \) denote the reference image, pose sequence, and speech audio, respectively. This formulation ensures that the model learns to effectively reverse the diffusion process, generating temporally coherent and identity-preserving animations conditioned on the given inputs.

% The training object can be formulated as,
% \begin{align}
% L_t = \mathbb{E}_{z_0,r,p,s} \left[ \left\|\epsilon - \epsilon_\theta(\tilde{z},r,p,s,t)\right\|^2 \right].
% \end{align}
% where $r,p,s$ stand for reference image, pose sequence and speech audio, respectively and $\epsilon_\theta$ represents the network parameter. Below, we detail the design of control mechanisms for each condition.

\noindent\textbf{Multi-human Pose Control.} 
% 1. Pose control for each human is independent.
% 2. Pose control should be idenity invariant.
% 3. number of individuals to control is unfixed(i.e. 2-4)
To address multi-human pose control, we propose the Multi-human Pose Encoder (MPE) as Pose Guider as shown in Figure~\ref{fig:baseline} (b). 
This module begins by utilizing instance masks to isolate individual human poses. 
Next, a shared convolutional network $\mathcal{F}_{\text{pose}}$ extracts features from each human pose $p_i$ separately.
Finally, these features are aggregated to generate a unified embedding that comprehensively represents the poses of all individuals:
% \begin{align}
%     e_{\text{pose}} = \sum_{i=1}^{n} e^i_{\text{pose}}
% \end{align}
%
\begin{align}
    e_{\text{pose}} = \sum_{i=1}^{n} e^i_{\text{pose}}; \text{ }e^i_{\text{pose}} = \mathcal{F}_{\text{pose}}(p_i), i = 1,2,...,n\text{,} 
\end{align}
where $e^i_{\text{pose}}\in\mathbb{R}^{f,c,w,h}$ stands for pose embedding of each human. This design is motivated by two key considerations. First, since pose control is independent for each individual, the model extracts and processes poses separately using a shared convolutional network, which promotes identity-invariant representations.  Second, given that the number of individuals is variable, our design enhances robustness by allowing the network to manage pose of each human independently rather than being confined to a fixed number of individual.

\noindent\textbf{Multi-human Identity Control.} 
% AnimateAnyone\cite{cui2024hallo2} employs ReferenceNet to maintain human identity, which has proven effective in preserving appearance. In its seeting, the reference image consistently centers on a single individual. However, 
As the reference image contains multiple identities needed to be controlled, we propose a Pose Adaptor with the MPE architecture that extracts multi-human spatial cues. 
First, the reference pose is input into the Pose Adaptor to obtain a pose embedding.
This embedding is then fused with the latent representation of the corresponding reference image and fed into ReferenceNet to provide spatial cues for each individual.
This approach effectively accommodates variations in both the positions and the number of individuals across cases.

\noindent\textbf{Multi-human Interactive Audio Control.} 
In split of the complex patterns of interaction mentioned in Section~\ref{sec:data}, the speaking scores of individual speakers serve as a good indicator of the underlying interaction patterns.
As shown in Figure~\ref{fig:baseline} (c), we proposed the Interactive Audio Driver(IAD) to model the alignment between audio features and the corresponding lip movements and facial expressions.
For $i$-th speaker, we use his speaking score $a^i\in\mathbb{R}^{f}$ to adjust the audio embedding $e_{\text{audio}}\in\mathbb{R}^{f,m,d}$. Subsequently, we employ the adjusted audio embedding \(e_{\text{audio}}^{i}\) and hidden features $h_k$ from the DenoisingNet to perform a cross-attention $\mathcal{F}_{\text{audio}}$ using a facial mask:
\begin{equation}
    h_{k+1} = h_k + \sum_{i=1}^{n} \mathcal{F}_{\text{audio}}(h_k, e_{\text{audio}}^i, \text{mask}_i) ;\text{ }e_{\text{audio}}^i = e_{\text{audio}} \cdot \sigma(\operatorname{MLP}(a^i))\text{,} 
\end{equation}
% \begin{equation}
%     h_{k+1} = h_k + \sum_{i=1}^{n} \mathcal{F}_{\text{audio}}(h_k, e_{\text{audio}}^i, \text{mask}_i) ,
% \end{equation}
\noindent where the parameters of this module are also shared across all speakers and $\text{mask}_i$ is obtained by the bounding box computed using three key head landmarks of human $i$.
This design not only ensures that the model learns an identity-invariant alignment between audio and facial features,
but also models the entire interactive process, thereby achieving a natural transition between listening and speaking.
As shown in Figure~\ref{fig:baseline} (a), the IAD module is inserted after each DenoisingNet block.

\begin{table*}[t]
\centering
\caption{\textbf{Quantitative Comparison and Ablation Study.} 
Experiments are conducted on the TonightShow for two-human scenarios and the LateLateShow for multi-human scenarios, under both easy and challenging test.
The data from TonightShow consists of conversations with $2$ speakers, while data from LateLateShow includes dialogues involving $2$ to $4$ speakers. Bold text indicates the best, while underlined text represents the second best.}
\setlength{\tabcolsep}{0.75mm} % Adjust column spacing
\begin{tabular}{l|ccc|ccc|ccc}
 \toprule
\multirow{2}{*}{Method} & 
\multicolumn{3}{c|}{Two Human} & 
\multicolumn{3}{c|}{Multiple Human} &
\multicolumn{3}{c}{All Test} \\
\cline{2-10}
& SSIM$\uparrow$ & PSNR$\uparrow$ & FVD$\downarrow$ 
& SSIM$\uparrow$ & PSNR$\uparrow$ & FVD$\downarrow$
& SSIM$\uparrow$ & PSNR$\uparrow$ & FVD$\downarrow$\\
\hline
\multicolumn{10}{l}{\textit{Comparison with Previous Methods}} \\
\hline
AnimateAnyone~\cite{hu2024animate} &  0.60 & 18.98 & 322.08
&0.64 &19.96    & 353.11 
&0.62 & \underline{19.47} & 337.60\\
ControlSVD~\cite{wu2024draganything} & 0.31 & 13.46 & 1036.96
& - & - & -
& - & - & -\\
CovOG & \textbf{0.62} & \textbf{19.16} & \textbf{306.01}
& \textbf{0.66} & \textbf{20.21} & \textbf{308.68}
& \textbf{0.64} & \textbf{19.69} & \textbf{307.35}\\
\hline
\multicolumn{10}{l}{\textit{Ablation Study}}  \\
\hline
CovOG w/o MPE & 0.60 & 18.88 & 317.41
& \underline{0.65} & \underline{20.00} & \underline{330.50}
& \underline{0.63} & 19.44 & \underline{323.96}\\
CovOG w/o IAD & \underline{0.61} & \underline{19.06} & \underline{313.69}
& \underline{0.65} & 19.86 & 347.92
& \underline{0.63} & 19.46 & 330.80\\

\bottomrule
\end{tabular}
\label{table:comparison}
\end{table*}
\section{Experiment}
\subsection{Datasets and Evaluation Metrics}
\textbf{Datasets.} 
% As shown in Table~\ref{tab:dataset_overview}, unlike existing datasets focus on single human generation such as SSGD~\cite{ginosar2019learning,zhang2021flow,zhou2022responsive,zhu2024infp}, MIT uniquely captures rich, multi-speaker full-body dynamics, setting a new benchmark for interactive speech-driven animation.
%
In our experiment, we first split the test set from the MIT datasets, which consists of approximately 200 easy cases and 200 challenging cases sourced from both the TonightShow and the LateLateShow.
The easy cases feature identities present in the training set but with novel pose and audio control parameters, whereas the challenging cases involve entirely unseen control signals to represent real application. 

\noindent\textbf{Evaluation Metrics.} 
To qualitatively analyze model performance, we utilize Structured Similarity (SSIM), Peak Signal-to-Noise Ratio (PSNR) and Frechet Inception Distance (FVD) to evaluate the quality of generated samples.
Unlike single-person talking head scenarios, lip alignment cannot be reliably assessed using LIPS~\cite{chung2017out} in our setting, as multi-person interactions involve both speaking and listening roles, often with side-facing views that LIPS is not designed to handle. How to effectively evaluate lip synchronization in such interactive contexts remains an open problem. To address this limitation, we complement our evaluation with user studies for visual-audio alignment.

\subsection{Implementation}
We pretrain our model following the two-stage paradigm proposed in AnimateAnyone~\cite{hu2024animate}, initializing it with weights from~\cite{Moore-AnimateAnyone}.
The model is trained on the entire training set, encompassing videos with varying numbers of speakers.
The first stage and the second stage all comprised $30,000$ steps with a resolution of 640$\times$384, frame number of $15$ and a batch size of $4$ on $4$ NVIDIA A6000 GPUs.
The Pose Adaptor is integrated into the first stage and remains fixed in the second stage, while Interactive Audio Driver is incorporated into the second stage with the motion module.
During inference, similar to Hallo2~\cite{cui2024hallo2}, we utilize the final six frames from the previous inference as motion frames, incorporating them as the initial six frames of the subsequent inference while keeping them fixed to ensure the continuity and smoothness of generation. In addition, we obtained audio embedding using Wav2Vec~\cite{baevski2020wav2vec}.

\subsection{Comparison}
\textbf{Quantitative Evaluation.}
We compare CovOG with two representative controllable video generation baselines: AnimateAnyone~\cite{hu2024animate} and ControlSVD~\cite{wu2024draganything}. While more recent methods have emerged~\cite{peng2024controlnext}, we select these two due to their simplicity and broad representativeness, which allow for clearer comparisons. To ensure fairness, AnimateAnyone follows the same inference setup as CovOG. For ControlSVD, we use pose embeddings as input to ControlNet, initialize from the first frame, and generate videos autoregressively.
As shown in Table~\ref{table:comparison}, CovOG consistently outperforms both baselines across all metrics.
AnimateAnyone struggles with multi-person scenarios, as its encoder jointly drives all subjects, while CovOG's MPE models each person independently and aggregates their effects.
Moreover, lacking audio control, AnimateAnyone produces random facial motions, whereas CovOG’s IAD leverages personalized audio embeddings to enhance head dynamics and ensure audio-visual alignment.
ControlSVD suffers from autoregressive error accumulation, leading to degraded quality over time, while CovOG maintains stability throughout generation.
\begin{wraptable}{r}{0.55\textwidth}  % r表示靠右放置，0.55可按需要调节宽度
\caption{\textbf{User Study}. ‘CC’, ‘BC’, and ‘AV-Align’ denote ‘character’, ‘background consistency’, and ‘audio-visual alignment’, respectively. ‘Visual’ indicates overall video quality.}
\small
\centering
\begin{tabular}{@{}lcccc@{}}
\toprule
Method & CC$\uparrow$ & BC$\uparrow$ & AV-Align$\uparrow$ & Visual$\uparrow$ \\ 
\midrule
\multicolumn{5}{l}{\textit{Comparison with Previous Methods}} \\
\midrule
AnimateAnyone~\cite{hu2024animate} & 2.81  & 3.83  & 2.66 & 2.64 \\
ControlSVD~\cite{wu2024draganything} & 2.57  & 1.86  & 1.86 & 1.57 \\
\textbf{CovOG} & \textbf{2.93}  & \textbf{4.11}  & \textbf{3.22} & \textbf{3.34} \\
\midrule
\multicolumn{5}{l}{\textit{Ablation Study}}  \\
\midrule
CovOG w/o MPE  & 2.64  & 3.55  &\underline{2.79} & 2.5\\
CovOG w/o IAD  & \underline{2.84}  &\underline{3.91}  &2.66  & \underline{2.81}\\
\bottomrule
\end{tabular}
\label{tab:user_study}
\end{wraptable}
\begin{wraptable}{p}{0.55\textwidth}
\caption{\textbf{Cross-modal Experiment}. ‘SC’, ‘BC’, ‘AQ’, and ‘IQ’ denote ‘subject consistency’, ‘background consistency’, ‘aesthetic quality’, and ‘imaging quality’, respectively.}
\small
\centering
\begin{tabular}{@{}lcccc@{}}
\toprule
Method & SC$\uparrow$ & BC$\uparrow$ & AQ$\uparrow$ & IQ$\uparrow$ \\ 
\midrule
AnimateAnyone~\cite{hu2024animate} & 0.945 & 0.952 & 0.530 & 0.564 \\
CovOG & \textbf{0.952} & \textbf{0.959} & \textbf{0.542} & \textbf{0.603} \\
\bottomrule
\end{tabular}
\label{tab:user_study_new}
\end{wraptable}
\noindent \textbf{User Study.}
We conduct a user study to evaluate character consistency, background consistency, audio-visual alignment, and overall visual quality.
Seven participants rated 10 randomly selected samples per method on a 1–5 scale(higher is better), based on the reference image and speaking score.
As shown in Table~\ref{tab:user_study}, CovOG outperforms other methods across all criteria, indicating superior control alignment and visual quality.

% \begin{table}[h!]
% \caption{\textbf{User Study}. `CC', `BC' and `AV-Align' stands for the `character', `background consistency' and `audio-visual alignment', respectively. 
% %
% `Visual' stands for overall video quality.}
% \centering
% \small
% \begin{tabular}{@{}lcccc@{}}
% \toprule
% Method & CC & BC & AV-Align & Visual \\ 
% \midrule
% \multicolumn{5}{l}{\textit{Comparison with Previous Methods}} \\
% \hline
% AnimateAnyone~\cite{hu2024animate} & 2.81  & 3.83  & 2.66 & 2.64 \\
% ControlSVD~\cite{wu2024draganything} & 2.57  & 1.86  & 1.86 & 1.57 \\
% CovOG & \textbf{2.93}  & \textbf{4.11}  & \textbf{3.22} & \textbf{3.34} \\
% \hline
% \multicolumn{5}{l}{\textit{Ablation Study}}  \\
% \hline
% CovOG w/o MPE  & 2.64  & 3.55  &\underline{2.79} & 2.5\\
% CovOG w/o IAD  & \underline{2.84}  &\underline{3.91}  &2.66  & \underline{2.81}\\
% \bottomrule
% \end{tabular}
% \label{tab:user_study}
% \end{table}
\noindent\textbf{Cross-modal Experiment.} To evaluate the generalization and practical applicability of our method, we conducted a cross-modal experiment. Specifically, we randomly selected 20 test cases by combining an identity image, a pose sequence, and corresponding speech audio from two different source videos, while ensuring that they involve the same number of speakers. Since ground-truth videos are unavailable for these cross-modal combinations, we employ VBench~\cite{huang2024vbench} to assess the generated results in terms of temporal consistency and visual quality, as shown in Table~\ref{tab:user_study_new}. The results demonstrate that CovOG achieves superior generalization both temporally and spatially.

\subsection{Ablation Study}
As shown in Table~\ref{table:comparison}, removing either MPE or IAD leads to a clear drop in performance across all metrics.
The absence of MPE results in the most significant decline, as torso control—essential for multi-person pose generation—heavily impacts visual quality.
Without IAD, the model lacks sufficient control signals, causing unnatural head movements due to the absence of audio guidance.
User study results in Table~\ref{tab:user_study} further confirm these findings: character and background consistency degrade without MPE, while audio-visual alignment suffers notably without IAD.
These results validate the complementary roles of MPE for multi-person pose control and IAD for audio-driven facial synchronization.

\subsection{Visualization Analysis}
\begin{figure*}[t]
    \centering
    \includegraphics[width=1\textwidth]{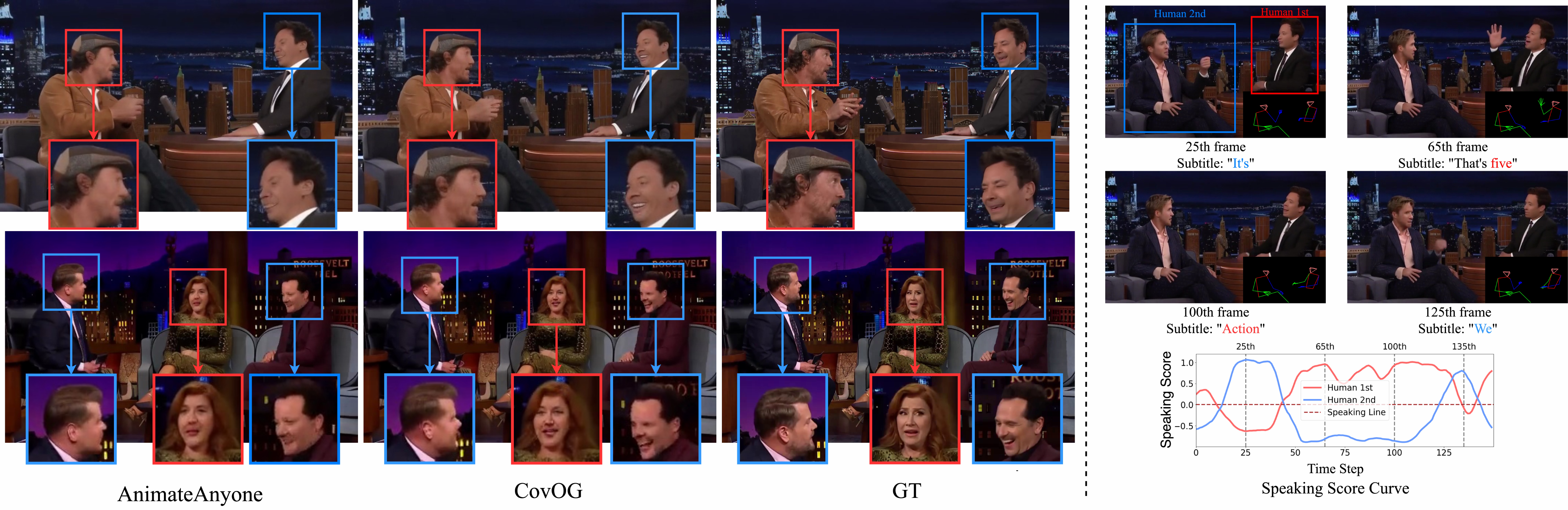}  
    \caption{\textbf{Qualitative Comparison} and \textbf{Interaction Visualization}. 
Left: The {\color{red}red} box indicates the speaker, and the {\color{blue}blue} box indicates the listener. 
Compared to AnimateAnyone, CovOG achieves superior lip synchronization for speakers and generates more natural, context-aware responses for listeners.
%
% AnimateAnyone often produces desynchronized lip movements and incoherent listener expressions.
%
Right: Visualization of the alignment with speaking scores, audio (\textit{i.e.}, subtitles), and pose.}

    \label{fig:vis}
\end{figure*}

\noindent\textbf{Qualitative Evaluation.}
We conduct qualitative evaluations on the MIT test set, as illustrated in Figure~\ref{fig:vis}, where the first row presents relatively simple cases and the second row includes more challenging ones. The red and blue bounding boxes indicate the speaker and listener, respectively. Both methods produce plausible gestures. However, AnimateAnyone tends to generate an \textbf{averaged face} for both speakers and listeners. For instance, the listener's mouth remains static, and the speaker exhibits only limited lip movement. In comparison, CovOG shows a higher degree of interactivity and closer alignment with the ground truth. The speaker appears more engaged in speech, while the listener displays responsive expressions such as laughter. This may be attributed to CovOG’s use of speaking scores to estimate speaking status, enabling adaptive facial expression generation. For example, when the input audio contains both speech and laughter, the model produces synchronized lip movements for the speaker and reactive expressions for the listener.

\noindent\textbf{Interaction Visualization.} We present the interaction visualization in the result generated by our CovOG, as shown in Figure~\ref{fig:vis}. 
The speaking score curve indicates a turn-taking dialogue between two individuals.
Key frames with their corresponding subtitles and the pose condition are displayed, with pronounced words highlighted in matching colors as in the speakings score curve. The results demonstrate that CovOG effectively aligns audio with lips and facial expressions for both speaker and listener, achieving natural interaction dynamics and strong audio-visual synchronization.

% \begin{figure}[t]
%     \centering
%     \includegraphics[width=0.47\textwidth]{sample.pdf} 
%     \caption{\textbf{Visualization of Interaction Generated by CovOG.} It demonstrates video alignment with speaking scores, speech audio (\textit{i.e.}, subtitles), and pose conditions.}
%     \label{fig:sample}
% \end{figure}

\subsection{Challenges in Multi-human Talking Scenarios}
Here, we outline the key challenges unique to multi-human talking scenarios in comparison to traditional talking-head and co-speech generation, and discuss the limitations of existing methods.

\noindent\textbf{Multi-huamn Interaction Modeling.} In a conversation, a person switches rapidly between speaking and listening, requiring the model to capture both the transitions and their dynamics.
During speaking, accurate lip–audio synchronization is crucial, while during listening, the model only needs to produce natural, context-appropriate reactions.
This difference in audio-visual patterns between speaking and listening poses a major challenge for generating realistic interactive speech.

\noindent\textbf{Side-Face Speech Alignment and Identity Consistency.}
In multi-person conversational scenarios, speakers frequently turn their heads to engage with others, resulting in side-face appearances during speech. Accurately modeling lip movements in such cases remains challenging, as most talking head generation methods are primarily optimized for frontal views~\cite{tan2024edtalk}. Furthermore, large rotational movements of the head and upper body pose challenges to maintaining visual consistency, particularly in facial features.

\noindent\textbf{Limitation of Existing Methods.} As discussed above, existing models face limitations in addressing these challenges. Moreover, talking-head methods are not designed to model full-body interactions, while co-speech models are often difficult to extend to multi-person scenarios. For instance, most recent work, TANGO~\cite{liu2024tango} requires a two-minute reference video to construct an interactive audio–frame graph, which is impractical in multi-person conversations where audio–frame pairs are sparse. This sparsity hinders the feasibility to retrieve keyframes, leading to performance degradation.

\section{Conclusion}
In this paper, we introduce the Multi-human Interactive Talking (MIT) dataset, the first large-scale benchmark for multi-person talking video generation. To demonstrate its utility, we propose CovOG, a baseline model that integrates pose and audio cues to generate natural multi-human talking videos. We hope this dataset fosters further research in more challenging human-centric video generation.

\bibliographystyle{plainnat}
\bibliography{main}  % 引用 main.bib

\begin{thebibliography}{60}
\providecommand{\natexlab}[1]{#1}
\providecommand{\url}[1]{\texttt{#1}}
\expandafter\ifx\csname urlstyle\endcsname\relax
  \providecommand{\doi}[1]{doi: #1}\else
  \providecommand{\doi}{doi: \begingroup \urlstyle{rm}\Url}\fi

\bibitem[Moo()]{Moore-AnimateAnyone}
Moore-animateanyone.
\newblock GitHub repository.
\newblock URL \url{https://github.com/MooreThreads/Moore-AnimateAnyone}.

\bibitem[Baevski et~al.(2020)Baevski, Zhou, Mohamed, and Auli]{baevski2020wav2vec}
Alexei Baevski, Yuhao Zhou, Abdelrahman Mohamed, and Michael Auli.
\newblock wav2vec 2.0: A framework for self-supervised learning of speech representations.
\newblock \emph{Advances in neural information processing systems}, 33:\penalty0 12449--12460, 2020.

\bibitem[Beliaev et~al.(2020)Beliaev, Rebryk, and Ginsburg]{beliaev2020talknet}
Stanislav Beliaev, Yurii Rebryk, and Boris Ginsburg.
\newblock Talknet: Fully-convolutional non-autoregressive speech synthesis model.
\newblock \emph{arXiv preprint arXiv:2005.05514}, 2020.

\bibitem[Blattmann et~al.(2023)Blattmann, Dockhorn, Kulal, Mendelevitch, Kilian, Lorenz, Levi, English, Voleti, Letts, et~al.]{blattmann2023stable}
Andreas Blattmann, Tim Dockhorn, Sumith Kulal, Daniel Mendelevitch, Maciej Kilian, Dominik Lorenz, Yam Levi, Zion English, Vikram Voleti, Adam Letts, et~al.
\newblock Stable video diffusion: Scaling latent video diffusion models to large datasets.
\newblock \emph{arXiv preprint arXiv:2311.15127}, 2023.

\bibitem[Chang et~al.(2023)Chang, Shi, Gao, Fu, Xu, Song, Yan, Yang, and Soleymani]{chang2023magicdance}
Di~Chang, Yichun Shi, Quankai Gao, Jessica Fu, Hongyi Xu, Guoxian Song, Qing Yan, Xiao Yang, and Mohammad Soleymani.
\newblock Magicdance: Realistic human dance video generation with motions \& facial expressions transfer.
\newblock \emph{CoRR}, 2023.

\bibitem[Chen et~al.(2023)Chen, Xia, He, Zhang, Cun, Yang, Xing, Liu, Chen, Wang, et~al.]{chen2023videocrafter1}
Haoxin Chen, Menghan Xia, Yingqing He, Yong Zhang, Xiaodong Cun, Shaoshu Yang, Jinbo Xing, Yaofang Liu, Qifeng Chen, Xintao Wang, et~al.
\newblock Videocrafter1: Open diffusion models for high-quality video generation.
\newblock \emph{arXiv preprint arXiv:2310.19512}, 2023.

\bibitem[Chen et~al.(2024)Chen, Cao, Chen, Li, and Ma]{chen2024echomimic}
Zhiyuan Chen, Jiajiong Cao, Zhiquan Chen, Yuming Li, and Chenguang Ma.
\newblock Echomimic: Lifelike audio-driven portrait animations through editable landmark conditions.
\newblock \emph{arXiv preprint arXiv:2407.08136}, 2024.

\bibitem[Chung and Zisserman(2017)]{chung2017out}
Joon~Son Chung and Andrew Zisserman.
\newblock Out of time: automated lip sync in the wild.
\newblock In \emph{Computer Vision--ACCV 2016 Workshops: ACCV 2016 International Workshops, Taipei, Taiwan, November 20-24, 2016, Revised Selected Papers, Part II 13}, pages 251--263. Springer, 2017.

\bibitem[Chung et~al.(2018)Chung, Nagrani, and Zisserman]{chung2018voxceleb2}
Joon~Son Chung, Arsha Nagrani, and Andrew Zisserman.
\newblock Voxceleb2: Deep speaker recognition.
\newblock \emph{arXiv preprint arXiv:1806.05622}, 2018.

\bibitem[Cui et~al.(2024)Cui, Li, Yao, Zhu, Shang, Cheng, Zhou, Zhu, and Wang]{cui2024hallo2}
Jiahao Cui, Hui Li, Yao Yao, Hao Zhu, Hanlin Shang, Kaihui Cheng, Hang Zhou, Siyu Zhu, and Jingdong Wang.
\newblock Hallo2: Long-duration and high-resolution audio-driven portrait image animation.
\newblock \emph{arXiv preprint arXiv:2410.07718}, 2024.

\bibitem[Feng et~al.(2023)Feng, Liu, Yu, Yao, Hui, Guo, Lin, Xue, Shi, Li, et~al.]{feng2023dreamoving}
Mengyang Feng, Jinlin Liu, Kai Yu, Yuan Yao, Zheng Hui, Xiefan Guo, Xianhui Lin, Haolan Xue, Chen Shi, Xiaowen Li, et~al.
\newblock Dreamoving: A human dance video generation framework based on diffusion models.
\newblock \emph{arXiv preprint arXiv:2312.05107}, 2023.

\bibitem[Geng et~al.(2023)Geng, Teotia, Tendulkar, Menon, and Vondrick]{geng2023affective}
Scott Geng, Revant Teotia, Purva Tendulkar, Sachit Menon, and Carl Vondrick.
\newblock Affective faces for goal-driven dyadic communication.
\newblock \emph{arXiv preprint arXiv:2301.10939}, 2023.

\bibitem[Ginosar et~al.(2019)Ginosar, Bar, Kohavi, Chan, Owens, and Malik]{ginosar2019learning}
Shiry Ginosar, Amir Bar, Gefen Kohavi, Caroline Chan, Andrew Owens, and Jitendra Malik.
\newblock Learning individual styles of conversational gesture.
\newblock In \emph{Proceedings of the IEEE/CVF Conference on Computer Vision and Pattern Recognition}, pages 3497--3506, 2019.

\bibitem[Guo et~al.(2023)Guo, Yang, Rao, Liang, Wang, Qiao, Agrawala, Lin, and Dai]{guo2023animatediff}
Yuwei Guo, Ceyuan Yang, Anyi Rao, Zhengyang Liang, Yaohui Wang, Yu~Qiao, Maneesh Agrawala, Dahua Lin, and Bo~Dai.
\newblock Animatediff: Animate your personalized text-to-image diffusion models without specific tuning.
\newblock \emph{arXiv preprint arXiv:2307.04725}, 2023.

\bibitem[He et~al.(2024)He, Huang, Zhang, Lin, Wu, Yang, Li, Chen, Xu, and Wu]{he2024co}
Xu~He, Qiaochu Huang, Zhensong Zhang, Zhiwei Lin, Zhiyong Wu, Sicheng Yang, Minglei Li, Zhiyi Chen, Songcen Xu, and Xiaofei Wu.
\newblock Co-speech gesture video generation via motion-decoupled diffusion model.
\newblock In \emph{Proceedings of the IEEE/CVF Conference on Computer Vision and Pattern Recognition}, pages 2263--2273, 2024.

\bibitem[Hu(2024)]{hu2024animate}
Li~Hu.
\newblock Animate anyone: Consistent and controllable image-to-video synthesis for character animation.
\newblock In \emph{Proceedings of the IEEE/CVF Conference on Computer Vision and Pattern Recognition}, pages 8153--8163, 2024.

\bibitem[Hu et~al.(2023)Hu, Lin, Gao, and Shou]{whisperV}
Siyuan Hu, Kevin~Qinghong Lin, Difei Gao, and Mike~Zheng Shou.
\newblock Whisperv.
\newblock GitHub repository, 2023.
\newblock URL \url{https://github.com/showlab/whisperV}.

\bibitem[Huang et~al.(2022)Huang, Huang, and Zhou]{huang2022perceptual}
Ailin Huang, Zhewei Huang, and Shuchang Zhou.
\newblock Perceptual conversational head generation with regularized driver and enhanced renderer.
\newblock In \emph{Proceedings of the 30th ACM international conference on multimedia}, pages 7050--7054, 2022.

\bibitem[Huang et~al.(2024)Huang, He, Yu, Zhang, Si, Jiang, Zhang, Wu, Jin, Chanpaisit, et~al.]{huang2024vbench}
Ziqi Huang, Yinan He, Jiashuo Yu, Fan Zhang, Chenyang Si, Yuming Jiang, Yuanhan Zhang, Tianxing Wu, Qingyang Jin, Nattapol Chanpaisit, et~al.
\newblock Vbench: Comprehensive benchmark suite for video generative models.
\newblock In \emph{Proceedings of the IEEE/CVF Conference on Computer Vision and Pattern Recognition}, pages 21807--21818, 2024.

\bibitem[Jiang et~al.(2024)Jiang, Liang, Yang, Lin, Zhong, and Zheng]{jiang2024loopy}
Jianwen Jiang, Chao Liang, Jiaqi Yang, Gaojie Lin, Tianyun Zhong, and Yanbo Zheng.
\newblock Loopy: Taming audio-driven portrait avatar with long-term motion dependency.
\newblock \emph{arXiv preprint arXiv:2409.02634}, 2024.

\bibitem[Jiang et~al.(2023)Jiang, Yang, Koh, Wu, Loy, and Liu]{jiang2023text2performer}
Yuming Jiang, Shuai Yang, Tong~Liang Koh, Wayne Wu, Chen~Change Loy, and Ziwei Liu.
\newblock Text2performer: Text-driven human video generation.
\newblock In \emph{Proceedings of the IEEE/CVF International Conference on Computer Vision}, pages 22747--22757, 2023.

\bibitem[Jin et~al.(2020)Jin, Xu, Xu, Wang, Liu, Qian, Ouyang, and Luo]{jin2020whole}
Sheng Jin, Lumin Xu, Jin Xu, Can Wang, Wentao Liu, Chen Qian, Wanli Ouyang, and Ping Luo.
\newblock Whole-body human pose estimation in the wild.
\newblock In \emph{Computer Vision--ECCV 2020: 16th European Conference, Glasgow, UK, August 23--28, 2020, Proceedings, Part IX 16}, pages 196--214. Springer, 2020.

\bibitem[Khirodkar et~al.(2024)Khirodkar, Bagautdinov, Martinez, Zhaoen, James, Selednik, Anderson, and Saito]{khirodkar2024sapiens}
Rawal Khirodkar, Timur Bagautdinov, Julieta Martinez, Su~Zhaoen, Austin James, Peter Selednik, Stuart Anderson, and Shunsuke Saito.
\newblock Sapiens: Foundation for human vision models.
\newblock In \emph{European Conference on Computer Vision}, pages 206--228. Springer, 2024.

\bibitem[Lei et~al.(2024)Lei, Wang, Ma, Huang, and Liu]{lei2024comprehensive}
Wentao Lei, Jinting Wang, Fengji Ma, Guanjie Huang, and Li~Liu.
\newblock A comprehensive survey on human video generation: Challenges, methods, and insights.
\newblock \emph{arXiv preprint arXiv:2407.08428}, 2024.

\bibitem[Li et~al.(2024)Li, Xu, Zhan, Mu, Li, Cheng, Chen, Chen, Ye, Wang, et~al.]{li2024openhumanvid}
Hui Li, Mingwang Xu, Yun Zhan, Shan Mu, Jiaye Li, Kaihui Cheng, Yuxuan Chen, Tan Chen, Mao Ye, Jingdong Wang, et~al.
\newblock Openhumanvid: A large-scale high-quality dataset for enhancing human-centric video generation.
\newblock \emph{arXiv preprint arXiv:2412.00115}, 2024.

\bibitem[Liu et~al.(2024{\natexlab{a}})Liu, Yang, Akiyama, Huang, Li, Kuriyama, and Taketomi]{liu2024tango}
Haiyang Liu, Xingchao Yang, Tomoya Akiyama, Yuantian Huang, Qiaoge Li, Shigeru Kuriyama, and Takafumi Taketomi.
\newblock Tango: Co-speech gesture video reenactment with hierarchical audio motion embedding and diffusion interpolation.
\newblock \emph{arXiv preprint arXiv:2410.04221}, 2024{\natexlab{a}}.

\bibitem[Liu et~al.(2024{\natexlab{b}})Liu, Guo, Zhen, Li, Ao, and Yan]{liu2024customlistener}
Xi~Liu, Ying Guo, Cheng Zhen, Tong Li, Yingying Ao, and Pengfei Yan.
\newblock Customlistener: Text-guided responsive interaction for user-friendly listening head generation.
\newblock In \emph{Proceedings of the IEEE/CVF Conference on Computer Vision and Pattern Recognition}, pages 2415--2424, 2024{\natexlab{b}}.

\bibitem[Liu et~al.(2022)Liu, Wu, Zhou, Xu, Qian, Lin, Zhou, Wu, Dai, and Zhou]{liu2022learning}
Xian Liu, Qianyi Wu, Hang Zhou, Yinghao Xu, Rui Qian, Xinyi Lin, Xiaowei Zhou, Wayne Wu, Bo~Dai, and Bolei Zhou.
\newblock Learning hierarchical cross-modal association for co-speech gesture generation.
\newblock In \emph{Proceedings of the IEEE/CVF Conference on Computer Vision and Pattern Recognition}, pages 10462--10472, 2022.

\bibitem[Ma et~al.(2024)Ma, He, Cun, Wang, Chen, Li, and Chen]{ma2024follow}
Yue Ma, Yingqing He, Xiaodong Cun, Xintao Wang, Siran Chen, Xiu Li, and Qifeng Chen.
\newblock Follow your pose: Pose-guided text-to-video generation using pose-free videos.
\newblock In \emph{Proceedings of the AAAI Conference on Artificial Intelligence}, volume~38, pages 4117--4125, 2024.

\bibitem[Ng et~al.(2022)Ng, Joo, Hu, Li, Darrell, Kanazawa, and Ginosar]{ng2022learning}
Evonne Ng, Hanbyul Joo, Liwen Hu, Hao Li, Trevor Darrell, Angjoo Kanazawa, and Shiry Ginosar.
\newblock Learning to listen: Modeling non-deterministic dyadic facial motion.
\newblock In \emph{Proceedings of the IEEE/CVF Conference on Computer Vision and Pattern Recognition}, pages 20395--20405, 2022.

\bibitem[Peebles and Xie(2023)]{peebles2023scalable}
William Peebles and Saining Xie.
\newblock Scalable diffusion models with transformers.
\newblock In \emph{Proceedings of the IEEE/CVF International Conference on Computer Vision}, pages 4195--4205, 2023.

\bibitem[Peng et~al.(2024)Peng, Wang, Zhang, Li, Yang, and Jia]{peng2024controlnext}
Bohao Peng, Jian Wang, Yuechen Zhang, Wenbo Li, Ming-Chang Yang, and Jiaya Jia.
\newblock Controlnext: Powerful and efficient control for image and video generation.
\newblock \emph{arXiv preprint arXiv:2408.06070}, 2024.

\bibitem[Prajwal et~al.(2020)Prajwal, Mukhopadhyay, Namboodiri, and Jawahar]{prajwal2020lip}
KR~Prajwal, Rudrabha Mukhopadhyay, Vinay~P Namboodiri, and CV~Jawahar.
\newblock A lip sync expert is all you need for speech to lip generation in the wild.
\newblock In \emph{Proceedings of the 28th ACM international conference on multimedia}, pages 484--492, 2020.

\bibitem[Qian et~al.(2021)Qian, Tu, Zhi, Liu, and Gao]{qian2021speech}
Shenhan Qian, Zhi Tu, Yihao Zhi, Wen Liu, and Shenghua Gao.
\newblock Speech drives templates: Co-speech gesture synthesis with learned templates.
\newblock In \emph{Proceedings of the IEEE/CVF International Conference on Computer Vision}, pages 11077--11086, 2021.

\bibitem[Rombach et~al.(2022)Rombach, Blattmann, Lorenz, Esser, and Ommer]{rombach2022high}
Robin Rombach, Andreas Blattmann, Dominik Lorenz, Patrick Esser, and Bj{\"o}rn Ommer.
\newblock High-resolution image synthesis with latent diffusion models.
\newblock In \emph{Proceedings of the IEEE/CVF conference on computer vision and pattern recognition}, pages 10684--10695, 2022.

\bibitem[Sepehri et~al.(2024)Sepehri, Fabian, Soltanolkotabi, and Soltanolkotabi]{sepehri2024mediconfusion}
Mohammad~Shahab Sepehri, Zalan Fabian, Maryam Soltanolkotabi, and Mahdi Soltanolkotabi.
\newblock Mediconfusion: Can you trust your ai radiologist? probing the reliability of multimodal medical foundation models.
\newblock \emph{arXiv preprint arXiv:2409.15477}, 2024.

\bibitem[Singer et~al.(2022)Singer, Polyak, Hayes, Yin, An, Zhang, Hu, Yang, Ashual, Gafni, et~al.]{singer2022make}
Uriel Singer, Adam Polyak, Thomas Hayes, Xi~Yin, Jie An, Songyang Zhang, Qiyuan Hu, Harry Yang, Oron Ashual, Oran Gafni, et~al.
\newblock Make-a-video: Text-to-video generation without text-video data.
\newblock \emph{arXiv preprint arXiv:2209.14792}, 2022.

\bibitem[Son~Chung et~al.(2017)Son~Chung, Senior, Vinyals, and Zisserman]{son2017lip}
Joon Son~Chung, Andrew Senior, Oriol Vinyals, and Andrew Zisserman.
\newblock Lip reading sentences in the wild.
\newblock In \emph{Proceedings of the IEEE conference on computer vision and pattern recognition}, pages 6447--6456, 2017.

\bibitem[Song et~al.(2024)Song, Hou, He, Ma, Wang, Sinha, Tsai, Luo, Dai, Chen, et~al.]{song2024directorllm}
Kunpeng Song, Tingbo Hou, Zecheng He, Haoyu Ma, Jialiang Wang, Animesh Sinha, Sam Tsai, Yaqiao Luo, Xiaoliang Dai, Li~Chen, et~al.
\newblock Directorllm for human-centric video generation.
\newblock \emph{arXiv preprint arXiv:2412.14484}, 2024.

\bibitem[Stypu{\l}kowski et~al.(2024)Stypu{\l}kowski, Vougioukas, He, Zi{\k{e}}ba, Petridis, and Pantic]{stypulkowski2024diffused}
Micha{\l} Stypu{\l}kowski, Konstantinos Vougioukas, Sen He, Maciej Zi{\k{e}}ba, Stavros Petridis, and Maja Pantic.
\newblock Diffused heads: Diffusion models beat gans on talking-face generation.
\newblock In \emph{Proceedings of the IEEE/CVF Winter Conference on Applications of Computer Vision}, pages 5091--5100, 2024.

\bibitem[Sung-Bin et~al.(2024)Sung-Bin, Chae-Yeon, Son, Hyun-Bin, Ju, Nam, and Oh]{sung2024multitalk}
Kim Sung-Bin, Lee Chae-Yeon, Gihun Son, Oh~Hyun-Bin, Janghoon Ju, Suekyeong Nam, and Tae-Hyun Oh.
\newblock Multitalk: Enhancing 3d talking head generation across languages with multilingual video dataset.
\newblock \emph{arXiv preprint arXiv:2406.14272}, 2024.

\bibitem[Tan et~al.(2024)Tan, Ji, Bi, and Pan]{tan2024edtalk}
Shuai Tan, Bin Ji, Mengxiao Bi, and Ye~Pan.
\newblock Edtalk: Efficient disentanglement for emotional talking head synthesis.
\newblock In \emph{European Conference on Computer Vision}, pages 398--416. Springer, 2024.

\bibitem[Tran et~al.(2024)Tran, Chang, Siniukov, and Soleymani]{tran2024dyadic}
Minh Tran, Di~Chang, Maksim Siniukov, and Mohammad Soleymani.
\newblock Dyadic interaction modeling for social behavior generation.
\newblock \emph{arXiv preprint arXiv:2403.09069}, 2024.

\bibitem[Vougioukas et~al.(2020)Vougioukas, Petridis, and Pantic]{vougioukas2020realistic}
Konstantinos Vougioukas, Stavros Petridis, and Maja Pantic.
\newblock Realistic speech-driven facial animation with gans.
\newblock \emph{International Journal of Computer Vision}, 128\penalty0 (5):\penalty0 1398--1413, 2020.

\bibitem[Wang et~al.(2023)Wang, Dai, Deng, and Wang]{wang2023agentavatar}
Duomin Wang, Bin Dai, Yu~Deng, and Baoyuan Wang.
\newblock Agentavatar: Disentangling planning, driving and rendering for photorealistic avatar agents.
\newblock \emph{arXiv preprint arXiv:2311.17465}, 2023.

\bibitem[Wang et~al.(2024)Wang, Weng, Li, Guo, Du, Niu, Ma, He, Wu, Hu, et~al.]{wang2024emotivetalk}
Haotian Wang, Yuzhe Weng, Yueyan Li, Zilu Guo, Jun Du, Shutong Niu, Jiefeng Ma, Shan He, Xiaoyan Wu, Qiming Hu, et~al.
\newblock Emotivetalk: Expressive talking head generation through audio information decoupling and emotional video diffusion.
\newblock \emph{arXiv preprint arXiv:2411.16726}, 2024.

\bibitem[Wang et~al.(2020)Wang, Wu, Song, Yang, Wu, Qian, He, Qiao, and Loy]{wang2020mead}
Kaisiyuan Wang, Qianyi Wu, Linsen Song, Zhuoqian Yang, Wayne Wu, Chen Qian, Ran He, Yu~Qiao, and Chen~Change Loy.
\newblock Mead: A large-scale audio-visual dataset for emotional talking-face generation.
\newblock In \emph{European Conference on Computer Vision}, pages 700--717. Springer, 2020.

\bibitem[Wu et~al.(2024)Wu, Li, Gu, Zhao, He, Zhang, Shou, Li, Gao, and Zhang]{wu2024draganything}
Weijia Wu, Zhuang Li, Yuchao Gu, Rui Zhao, Yefei He, David~Junhao Zhang, Mike~Zheng Shou, Yan Li, Tingting Gao, and Di~Zhang.
\newblock Draganything: Motion control for anything using entity representation.
\newblock In \emph{European Conference on Computer Vision}, pages 331--348, 2024.

\bibitem[Xu et~al.(2024)Xu, Zhang, Liew, Yan, Liu, Zhang, Feng, and Shou]{xu2024magicanimate}
Zhongcong Xu, Jianfeng Zhang, Jun~Hao Liew, Hanshu Yan, Jia-Wei Liu, Chenxu Zhang, Jiashi Feng, and Mike~Zheng Shou.
\newblock Magicanimate: Temporally consistent human image animation using diffusion model.
\newblock In \emph{Proceedings of the IEEE/CVF Conference on Computer Vision and Pattern Recognition}, pages 1481--1490, 2024.

\bibitem[Yang et~al.(2024)Yang, Teng, Zheng, Ding, Huang, Xu, Yang, Hong, Zhang, Feng, et~al.]{yang2024cogvideox}
Zhuoyi Yang, Jiayan Teng, Wendi Zheng, Ming Ding, Shiyu Huang, Jiazheng Xu, Yuanming Yang, Wenyi Hong, Xiaohan Zhang, Guanyu Feng, et~al.
\newblock Cogvideox: Text-to-video diffusion models with an expert transformer.
\newblock \emph{CoRR}, 2024.

\bibitem[Zhang et~al.(2024)Zhang, Wu, Liu, Zhao, Ran, Gu, Gao, and Shou]{zhang2024show}
David~Junhao Zhang, Jay~Zhangjie Wu, Jia-Wei Liu, Rui Zhao, Lingmin Ran, Yuchao Gu, Difei Gao, and Mike~Zheng Shou.
\newblock Show-1: Marrying pixel and latent diffusion models for text-to-video generation.
\newblock \emph{International Journal of Computer Vision}, pages 1--15, 2024.

\bibitem[Zhang et~al.(2023{\natexlab{a}})Zhang, Rao, and Agrawala]{zhang2023adding}
Lvmin Zhang, Anyi Rao, and Maneesh Agrawala.
\newblock Adding conditional control to text-to-image diffusion models.
\newblock In \emph{Proceedings of the IEEE/CVF International Conference on Computer Vision}, pages 3836--3847, 2023{\natexlab{a}}.

\bibitem[Zhang et~al.(2023{\natexlab{b}})Zhang, Cun, Wang, Zhang, Shen, Guo, Shan, and Wang]{zhang2023sadtalker}
Wenxuan Zhang, Xiaodong Cun, Xuan Wang, Yong Zhang, Xi~Shen, Yu~Guo, Ying Shan, and Fei Wang.
\newblock Sadtalker: Learning realistic 3d motion coefficients for stylized audio-driven single image talking face animation.
\newblock In \emph{Proceedings of the IEEE/CVF Conference on Computer Vision and Pattern Recognition}, pages 8652--8661, 2023{\natexlab{b}}.

\bibitem[Zhang et~al.(2021)Zhang, Li, Ding, and Fan]{zhang2021flow}
Zhimeng Zhang, Lincheng Li, Yu~Ding, and Changjie Fan.
\newblock Flow-guided one-shot talking face generation with a high-resolution audio-visual dataset.
\newblock In \emph{Proceedings of the IEEE/CVF Conference on Computer Vision and Pattern Recognition}, pages 3661--3670, 2021.

\bibitem[Zhou et~al.(2022{\natexlab{a}})Zhou, Bai, Zhang, Yao, Zhao, and Mei]{zhou2022responsive}
Mohan Zhou, Yalong Bai, Wei Zhang, Ting Yao, Tiejun Zhao, and Tao Mei.
\newblock Responsive listening head generation: a benchmark dataset and baseline.
\newblock In \emph{European Conference on Computer Vision}, pages 124--142. Springer, 2022{\natexlab{a}}.

\bibitem[Zhou et~al.(2023)Zhou, Bai, Zhang, Yao, and Zhao]{zhou2023interactive}
Mohan Zhou, Yalong Bai, Wei Zhang, Ting Yao, and Tiejun Zhao.
\newblock Interactive conversational head generation.
\newblock \emph{arXiv preprint arXiv:2307.02090}, 2023.

\bibitem[Zhou et~al.(2022{\natexlab{b}})Zhou, Yang, Li, Saito, Aneja, and Kalogerakis]{zhou2022audio}
Yang Zhou, Jimei Yang, Dingzeyu Li, Jun Saito, Deepali Aneja, and Evangelos Kalogerakis.
\newblock Audio-driven neural gesture reenactment with video motion graphs.
\newblock In \emph{Proceedings of the IEEE/CVF conference on computer vision and pattern recognition}, pages 3418--3428, 2022{\natexlab{b}}.

\bibitem[Zhu et~al.(2022)Zhu, Wu, Zhu, Jiang, Tang, Zhang, Liu, and Loy]{zhu2022celebv}
Hao Zhu, Wayne Wu, Wentao Zhu, Liming Jiang, Siwei Tang, Li~Zhang, Ziwei Liu, and Chen~Change Loy.
\newblock Celebv-hq: A large-scale video facial attributes dataset.
\newblock In \emph{European conference on computer vision}, pages 650--667. Springer, 2022.

\bibitem[Zhu et~al.(2023)Zhu, Liu, Liu, Qian, Liu, and Yu]{zhu2023taming}
Lingting Zhu, Xian Liu, Xuanyu Liu, Rui Qian, Ziwei Liu, and Lequan Yu.
\newblock Taming diffusion models for audio-driven co-speech gesture generation.
\newblock In \emph{Proceedings of the IEEE/CVF Conference on Computer Vision and Pattern Recognition}, pages 10544--10553, 2023.

\bibitem[Zhu et~al.(2024)Zhu, Zhang, Rong, Hu, Liang, and Ge]{zhu2024infp}
Yongming Zhu, Longhao Zhang, Zhengkun Rong, Tianshu Hu, Shuang Liang, and Zhipeng Ge.
\newblock Infp: Audio-driven interactive head generation in dyadic conversations.
\newblock \emph{arXiv preprint arXiv:2412.04037}, 2024.

\end{thebibliography}

\end{document}